\documentclass{article}

\PassOptionsToPackage{numbers}{natbib}

    \usepackage[preprint]{neurips_data_2023}

\usepackage[utf8]{inputenc} 
\usepackage[T1]{fontenc}    
\usepackage{hyperref}       
\usepackage{url}            
\usepackage{booktabs}       
\usepackage{amsfonts}       
\usepackage{nicefrac}       
\usepackage{microtype}      
\usepackage{xcolor}         

\usepackage{caption}
\usepackage{graphicx}
\usepackage{enumitem} 
\usepackage{xspace}
\usepackage{amsmath}
\usepackage{amssymb}
\usepackage{multirow}
\usepackage{algorithm}
\usepackage{algorithmic}
\usepackage{wrapfig}
\usepackage{subcaption}
\usepackage{bm}
\usepackage{listings}

\usepackage{lipsum}

\usepackage{color}

\usepackage{longtable}
\usepackage{stackengine}
\newcommand\blivet[2]{\stackengine{-.1ex}{#1}{\stackon[.5pt]{\CAL}{#2}}{O}{c}{F}{T}{L}}
\newcommand\CAL{\scalebox{1.7}{\rotatebox[origin=center]{170}{$\circlearrowleft$}}}
\stackMath

\title{VISION Datasets: A Benchmark for \underline{V}ision-based \underline{I}ndu\underline{S}trial \underline{I}nspecti\underline{ON}}

\author{
    Haoping Bai\thanks{Corresponding author: Haoping Bai \texttt{haoping\_bai@apple.com}.} \quad Tatiana Likhomanenko \quad Oncel Tuzel \\
    \textbf{Ping Huang \quad Jiulong Shan \quad Meng Cao} \\
    Apple \\
    \\
    \textbf{Shancong Mou \quad Jianjun Shi} \\
    Georgia Institute of Technology \\
    \\
    \textbf{Gokberk Cinbis} \\
    Middle East Technical University \\
}

\begin{document}

\maketitle

\begin{abstract} \label{sec:abstract}

Despite progress in vision-based inspection algorithms, real-world industrial challenges -- specifically in data availability, quality, and complex production requirements -- often remain under-addressed. We introduce the VISION Datasets, a diverse collection of 14 industrial inspection datasets, uniquely poised to meet these challenges. Unlike previous datasets, VISION brings versatility to defect detection, offering annotation masks across all splits and catering to various detection methodologies. Our datasets also feature instance-segmentation annotation, enabling precise defect identification. With a total of 18k images encompassing 44 defect types, VISION strives to mirror a wide range of real-world production scenarios. By supporting two ongoing challenge competitions on the VISION Datasets, we hope to foster further advancements in vision-based industrial inspection. The datasets are available at \url{https://huggingface.co/datasets/VISION-Workshop/VISION-Datasets}.

\end{abstract}
\section{Introduction}\label{sec:intro}

\begin{figure}[!ht]
    \centering
    \vspace{-\belowcaptionskip} 
    \vspace{-\textfloatsep}     
    \includegraphics[width=\textwidth,height=\textheight,keepaspectratio]{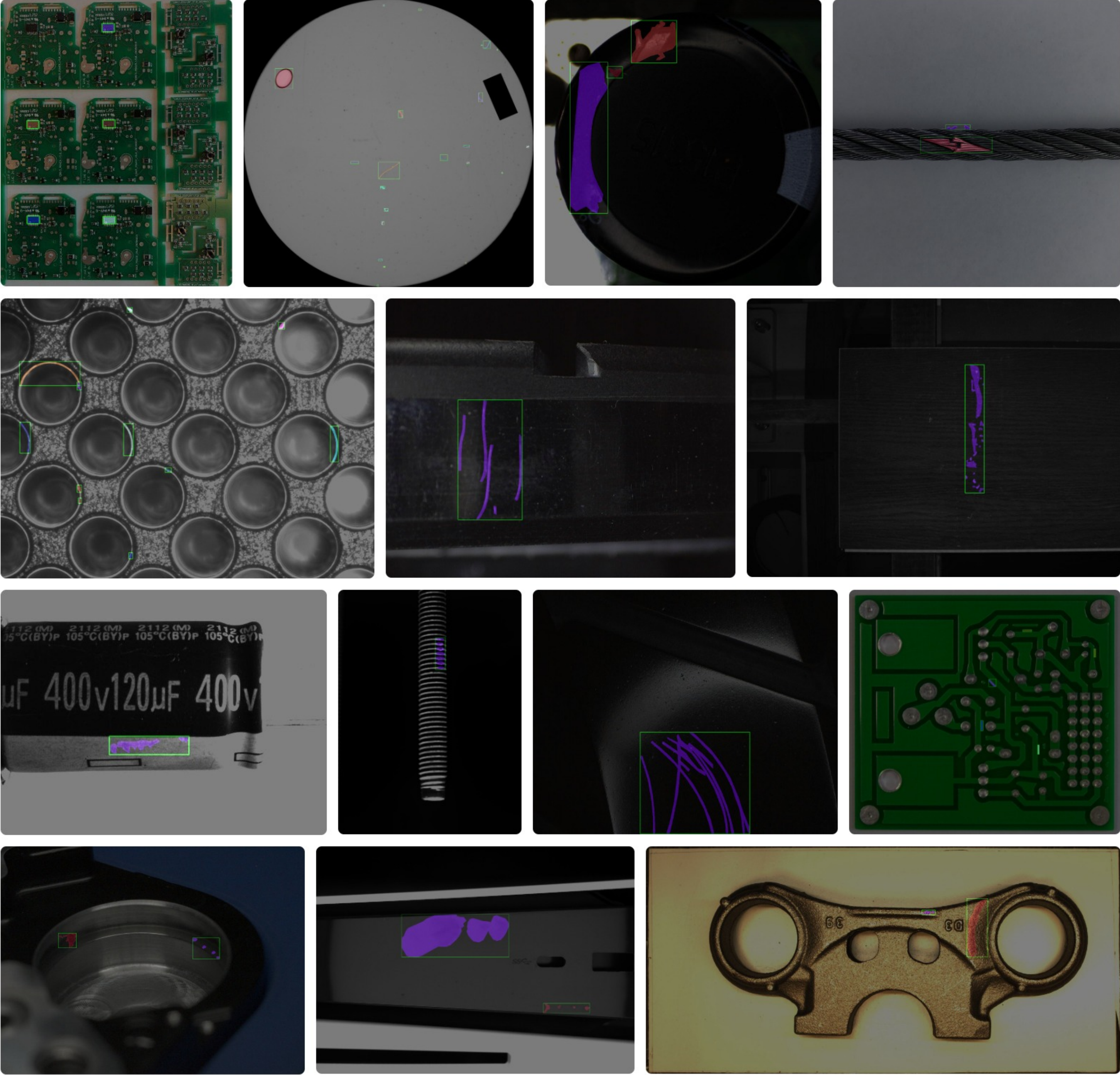}
    \caption{Overview of the VISION Datasets.}
    \label{fig:vision_datasets}
\end{figure}

Recent years have witnessed the proliferation of vision-based anomaly detection algorithms that have shown promising performance across a range of publicly available datasets \cite{bhatt2021image,chalapathy2019deep,pang2021deep,cui2022survey,mou2022synthetic}. However, these advancements, though impressive, are often found wanting in real-world industrial scenarios, owing to unique challenges that often go unrepresented in existing datasets. These challenges primarily revolve around the trifecta of data availability, data quality, and the complexity of production requirements, which can vary considerably across diverse industrial settings.

Consequently, the urgent need arises for a more robust and comprehensive dataset that encapsulates these complexities and lays a solid foundation for the development and evaluation of vision-based inspection algorithms. In response to this need, we present the VISION Datasets --- a collection of 14 industrial inspection datasets that cover a wide array of manufacturing processes, materials, and industries. Designed to navigate the intricacies of vision-based industrial inspection, the VISION Datasets offer a unique benchmarking opportunity for researchers and practitioners alike.

The principal contributions of the VISION Datasets are as follows:

\begin{enumerate}
    \item \textbf{Versatility}: Unlike previous datasets \cite{bergmann2019mvtec, mishra21-vt-adl} that solely focus on the self-supervised setting, the VISION Datasets provide the necessary flexibility to benchmark a variety of detection algorithms, including unsupervised, weakly-supervised, semi-supervised, and supervised defect detection methods. This flexibility is provided by the inclusion of annotation masks across all training, validation, and testing splits.
    \item \textbf{Rich annotation}: Unlike other datasets that feature semantic segmentation style annotation masks, the VISION Datasets offer instance-segmentation style labeling. This detailed annotation approach distinguishes between defects of the same type within the same image, facilitating training and evaluation of models capable of identifying individual defects.
    \item \textbf{Scale and diversity}: With 18k images encompassing 44 distinct defect types, the VISION Datasets cover a broad range of manufacturing processes, materials, and industries. This scale and diversity make the datasets an apt representative of real-world production scenarios.
    \item \textbf{Catalyst for innovation}: The VISION Datasets are accompanied by two challenge competitions that call together researchers and practitioners from both academia and industry. The combination of datasets and competitions serves as a catalyst for the development of innovative solutions to tackle the unique challenges represented by the datasets.
\end{enumerate}

By aiming to capture the diverse challenges inherent in real-world complexities, the VISION Datasets seek to bridge the gap between academic research and industrial practice, thereby driving further advancements in vision-based industrial inspection systems.

\section{Background}

\subsection{Related Datasets and Benchmarks}
While a review of currently available industrial inspection datasets can be found in \cite{cui2022survey}, in Table \ref{table:datasets}  we provide a comparison of the VISION Datasets with existing vision-based inspection datasets with segmentation annotation.

\renewcommand{\arraystretch}{1.2} 
\begin{table}[!t]
\centering
\begin{tabular}{l|cccccc}
\hline
\multicolumn{1}{l|}{\begin{tabular}[c]{@{}c@{}}Dataset\end{tabular}} & \multicolumn{1}{c}{\begin{tabular}[c]{@{}c@{}}Images\end{tabular}} & \multicolumn{1}{c}{\begin{tabular}[c]{@{}c@{}}Annotated\\ Images\end{tabular}} & \multicolumn{1}{c}{\begin{tabular}[c]{@{}c@{}}Annotation\\ Masks\end{tabular}} & \multicolumn{1}{c}{\begin{tabular}[c]{@{}c@{}}Instance\\ Label\end{tabular}} & \multicolumn{1}{c}{\begin{tabular}[c]{@{}c@{}}Products\end{tabular}} & \multicolumn{1}{c}{\begin{tabular}[c]{@{}c@{}}Defect\\ Types\textsuperscript{\textdagger} \end{tabular}} \\ \hline
AITEX* \cite{aitex} & 205 & 105 & 105 &  & 7* & 12 \\ \hline
BeanTech \cite{mishra21-vt-adl} & 1799 & 290 & 290 &  & 3 & 3 \\ \hline
KolektorSDD \cite{Tabernik2019JIM} & 399 & 52 & 52 &  & 1 & 1 \\ \hline
KolektorSDD2 \cite{Bozic2021COMIND} & 3335 & 356 & 356 &  & 1 & 1 \\ \hline
MTD \cite{8560423} & 1344 & 392 & 392 &  & 1 & 5 \\ \hline
MVTec AD \cite{bergmann2019mvtec} & 5354 & 1258 & 1258 &  & 15 & 69 \\ \hline
VISION V1\textsuperscript{\dag}  & 18422 & 4165 & 9837 & \checkmark & 14 & 44 \\ \hline
VISION V2\textsuperscript{\ddag}  & 18422 & 13804 & 38272 & \checkmark & 14 & 44 \\ \hline
\end{tabular}
\vspace{5px}
\caption{Comparison with Real World Manufacturing Datasets with Segmentation Annotation. VISION Datasets are the largest and one of the most diverse manufacturing benchmark datasets to date in terms of both image number and available annotations. For dataset with binary defect label that does not distinguish among defect types, we count one defect type for each product in the dataset. *AITEX provides data on 7 types of fabric products instead of 7 distinctive product types.  \textsuperscript{\dag}VISION V1 refers to the currently released version, which is also used in the challenge competitions. \textsuperscript{\ddag}VISION V2 refers to the full dataset that we plan to make available through future challenges.}
\label{table:datasets}
\end{table}

Among available vision-based inspection AITEX, KoektorSDD, KoektorSDD2, MTD, all focus on a single product. While BeanTech and MVTec AD feature multiple products, the two datasets are mainly designed for benchmarking unsupervised algorithms. Both datasets contain no defective samples in the training set and only feature a limited amount of annotation masks in the testing split for evaluation purpose. VISION Datasets, on the other hand, provide annotation masks across training, validation, and testing splits to allow the full flexibility to benchmark unsupervised, weakly-supervised, semi-supervised, and supervised defect detection algorithms.

In addition, all the previously existing datasets are annotated with semantic segmentation style annotation masks, where defects of the same type are not distinguished when occurring in the same image. While datasets that often feature one single defect per image can technically be extended to the instance segmentation setting, they lack the data for supervision, and more importantly, for evaluation of model capability on handling images with multiple and potentially overlapping defects. In contrast, the instance-segmentation style labeling provides VISION Datasets the unique capability to train and evaluate models on identifying individual defect and gauging the corresponding specifications.

\section{The VISION Datasets}

\subsection{Overview}
The VISION Datasets is a collection of 14 industrial inspection datasets sourced from Roboflow, designed to explore the unique challenges of vision-based industrial inspection. These datasets cover a wide range of manufacturing processes, materials, and industries. The datasets have a total of 18k images. To support two ongoing challenge competitions hosted under our affiliated workshop, we provide the VISION V1 dataset with 10k high-quality defect segmentation annotations on 4k images, spanning 44 defect types. The remaining 14k images are released without labels to simulate the data-rich and annotation-rare scenario in real-world production and to encourage the development of effective self-supervised and semi-supervised algorithms. The VISION V2 dataset with full annotation will be released as a part of a subsequent challenge following the conclusion of our workshop.

\subsection{Curation Process and Rationale}

Crowd-sourcing platforms for data collection offer unparalleled opportunities for gathering diverse datasets to assess the generalization capability of algorithms \cite{li2022elevater, ciaglia2022roboflow}. Yet, this approach is not without its challenges. When sourcing data from a multitude of contributors, consistency, accuracy, and quality control can become significant concerns. These platforms frequently lack rigorous checks to ensure that the data being collected is of the highest quality. This may result in datasets that contain erroneous entries, duplicates, and misrepresentations. Even more concerning is the potential for inadvertent data leakage across training, validation, and test sets, which could lead to overestimated performance and unrealistic expectations for machine learning models. 

To ensure the quality and relevance of sourced datasets, we assembled a team of researchers, engineers, and annotators with extensive experience in working with vision-based inspection datasets and screened over 1800 manufacturing-related datasets from Roboflow. The search produced a ranked list of datasets based on relevance to defect detection and annotation quality. The top 14 ranking datasets are selected based on the consensus that the datasets reflect realistic production challenges.

Due to both the consideration to remain faithful to naturally existing label challenges and the difficulty in distinguishing between unintentional labeling oversight and domain-specific judgments without the manufacturers' specification sheets, we refrain from modifying original defect decisions. To enable precise defect detection even with existing label limitations, we provide refined segmentation masks for each defect indicated by the original bounding boxes.

\subsection{Annotating Defect Instance Segmentation Labels}
We leverage an annotation team with extensive prior experience in producing high-quality annotations for manufacturing-related datasets. The annotators practiced on a subset of each dataset until reaching a satisfactory level of quality and inter-rater reliability before the full annotation. Any defect with an ambiguous boundary will be flagged for verification by other annotators until a consensus is established. The annotation effort spans a total of 3200 working hours.


\subsection{Building Dataset Splits}

To ensure the benchmark can faithfully reflect the performance of algorithms, we need to minimize leakage across train, validation, and testing data. Due to the crowd-sourced nature, the original dataset splits are not always guaranteed to be free of leakage. As a result, we design a process to resplit the datasets with specific considerations for industrial defect detection.

Given distinct characteristics of defect detection datasets, including but not limited to:

\begin{itemize}
    \item Stark contrast between large image size and small defect size
    \item Highly aligned non-defective images may seem to be duplicates, but are necessary to represent natural distribution and variation to properly assess the false detection rate. 
\end{itemize}

Naively deduplicating with image-level embedding or hash would easily drown out small defects and regard distinct non-defective images as duplicates. Therefore, we first only deduplicate images with identical byte contents and set the images without defect annotation aside. For images with defect annotations, we want to reduce leakage at the defect level. Therefore, we follow \cite{bai2021selfsupervised} and train a self-supervised similarity model on the defect regions and model the similarity between two images as the maximum pairwise similarity between the defects on each image. To further follow production practice, we manually identify datasets that record product serial number in the file name, and assign images of the same unit to the same connected compenents to prevent leakage at the unit level. Finally, we perform connected component analysis on the image similarity graph. 

To limit the number of available samples, we randomly keep at most 200 connected components for each defect type. Then we randomly assign components to train, validation, and testing split with maximum capacity at 30, 30, and 140, respectively. Whenever a component is assigned to a split that has reached the designated capacity, the component is reassigned to another random split. Note that for datasets with a high number of defects per image, keeping enough samples for some rare defect types will lead to additional samples for other defects. The detailed dataset information can be found in the Appendix \ref{sec:appendix}.

To discourage manual exploitation during the data competition, the discarded images are provided alongside the test split data as the inference data for participants to generate their submissions. However, the testing performance is evaluated exclusively based on the test split data. The exact testing image ids and the dataset-splitting script will be released after the competition concludes.

\section{VISION Challenges}

We host two challenge tracks, each addressing the challenges in the industrial inspection domain from a unique angle. Track 1\footnote{\url{https://bit.ly/VISION_Track_1}} takes a model-centric approach, while Track 2\footnote{\url{https://bit.ly/VISION_Track_2}} adopts a data-centric perspective:

\vspace{-.5cm}
\begin{align*}
\text{Track 1: }& \text{Data} \rightarrow \blivet{\text{Model}}{} & \text{Track 2: }& \blivet{\text{Data}}{}\rightarrow \text{Model}
\end{align*}

The competitions are designed to push the boundaries of what's currently possible in the field of machine learning for industrial defect detection. By fostering a competitive environment where different teams put their algorithms to the test, we hope to spur innovation and progress in this critical area of application. Ultimately, the goal is to yield machine learning solutions that can be effectively and efficiently implemented in real-world manufacturing processes, contributing to increased product quality and reduced waste.

At the time of the paper, the Track 1 competition has attracted over 30 registered teams representing 12 companies and 20 universities, while the Track 2 competition features participation from 11 companies and 10 universities.

\subsection{Track 1: Data-Efficient Defect Detection}

The primary goal of this competition is to assess the data efficiency of machine learning algorithms. This stems from the realization that real-world manufacturing processes often come with numerous realistic data and annotation constraints. In such scenarios, it is not uncommon to have a limited volume of training data and annotations, which may not be sufficient to adequately train sophisticated machine learning models from scratch. Thus, accurately detecting defects with a limited amount of data is highly desirable for a machine learning model in this context, and the limited training data provided with the VISION Datasets can effectively test the data efficiency of the participating models.

Another main challenge in the track 1 competition is the diversity of products and surface inspection tasks. In the real-world manufacturing sector, the range of products that require quality inspection is vast. Each product, with its unique surface characteristics, requires a tailored inspection approach. Moreover, the types of defects that can occur vary greatly, from scratches and dents to discoloration and deformations. The robustness of an algorithm's ability to adapt to this diversity and still maintain a high level of accuracy is another key factor this competition aims to evaluate.

\textbf{Evaluation Metrics:} In addition to the standard mAP (mean Average Precision) metric used in instance segmentation competitions, we adopted a composite metric for the Track 1 competition. The evaluation metric is a weighted average of mAP and mAR (mean Average Recall): $0.5 * \text{mAP} + 0.5 * \text{mAR}^{\text{max}=100}$. The final metric will be the average across all 14 datasets, each bearing equal weight.

The mAP metric is a popular metric used in object detection. It provides a single-figure measure of the quality of the predictions across all recall levels. It essentially quantifies how well the model identifies positive instances throughout the ranked list of predictions.

The $\text{mAR}^{\text{max}=100}$ metric, on the other hand, measures the ability of the model to identify as many positive instances as possible within the top 100 ranked predictions. It is a metric that pays particular attention to recall, emphasizing the need to catch as many true positives as possible. Though the mAP calculation inherently considers recall, integrating mAR further underscores the importance of accurately identifying and ranking all defects. This approach more closely aligns with real-world scenarios where companies prioritize the prevention of defective products from entering the market.

The combination of these two metrics, mAP and $\text{mAR}^{\text{max}=100}$, is specifically designed for manufacturing applications where both overkills (false alarms) and escapes (missed defects) can have significant consequences. False alarms, indicated by a drop in precision, can lead to unnecessary checks or halts in the production line, leading to inefficiency and cost overrun. Missed defects, signaled by a decrease in the recall, can result in defective products reaching the market, causing potential harm to brand reputation and incurring additional costs associated with returns and repairs.

The averaging across all 14 datasets with equal weights, meanwhile, ensures that the models are well-rounded and robust, capable of handling a diverse range of products and defects. This is consistent with the challenge's goal of seeking solutions that can perform well across a wide range of real-world manufacturing contexts.

\subsection{Track 2: Data-Generation for Defect Detection}

In the second challenge, our focus shifts from data-efficient defect detection to data-generation for defect detection. This shift is driven by numerous challenges that are commonly encountered in high-stake manufacturing scenarios, many of which can be attributed to data limitations.

In real-world manufacturing, there is often an extreme imbalance between normal and defective samples in the data. This is because defects, by their very nature, are anomalies and do not occur as frequently as non-defective instances. This imbalance can make it challenging for machine learning models to learn effectively. When the model is trained on a dataset where one class vastly outnumbers the other, it can become biased towards predicting the majority class, resulting in poor performance in identifying the minority class, i.e., the defects.

Furthermore, there is the issue of `long-tailed' defects. These are defect types that occur infrequently and thus provide insufficient examples for supervised model training. Despite their rarity, these long-tailed defects are costly to ignore as they can lead to serious consequences, such as product failures or safety issues.

Given these challenges, it becomes clear that developing techniques that can directly improve the quality and quantity of available data is equally important as building high-performance machine learning models. The more diverse and representative the training data, the better equipped the model will be to handle the wide array of defect types it may encounter in a real-world setting.

In this challenge competition, we provide a fixed model and initial dataset for the vision-based inspection task. The focus here will not be on the model itself, but on the data. Participants are encouraged to apply data cleaning and augmentation techniques to enhance the existing data. Data cleaning refers to the process of detecting and correcting or removing corrupt, inaccurate, or irrelevant parts of the data, thereby improving its quality. Data augmentation, on the other hand, involves techniques that increase the amount of training data using information available in the existing data. This could include methods like flipping, rotating, zooming, or cropping images in a dataset.

Moreover, participants are urged to generate synthetic data based on the provided data. This could be accomplished using generative models, which are capable of creating new data instances that resemble the training data, or rendering engines, which can produce realistic images of objects from 3D models.

Importantly, the participants will only submit the altered datasets for evaluation. The goal is not to improve the defect detection model, but to create the most useful dataset for the task. The challenge, thus, seeks to stimulate innovation in data enhancement techniques, which can significantly impact the effectiveness of machine learning models in defect detection.

\section{Insights from Competition Winners}

To understand effective strategies for defect detection in industrial inspection, we reviewed the technical reports shared by the top-performing teams in each challenge track. Here, we share a summary of the winning solutions and common insights.

\subsection{Track 1}

\begin{description}
    \item[1st Place:] The top team used an ensemble of two state-of-the-art object detection models, EVA \cite{EVA} and MaskDino \cite{li2022mask}, adding a Dice loss to the original cross entropy (CE) loss. They trained their model on all 14 datasets simultaneously and reported improved performance than training on each dataset separately.
    \item[2nd Place:] The second-place team used a multi-stage solution with a HTC \cite{Chen2019HybridTC} detection architecture, with a backbone of two Swin-B \cite{liu2021Swin} models composed via CBNetV2 \cite{9932281}. The outputs were further refined with a Mask2Former \cite{cheng2021mask2former} with Swin-L backbone and other models to improve recall. They also applied Test Time Augmentation (TTA) during inference.
    \item[3rd Place:] The third-place team used an ensemble of 28 models, each trained on the combination of all 14 datasets. This ensemble included the Cascade-RCNN \cite{cai2018cascade} detectors with varying backbones including ResNet50, Swin model family, and composite backbones from CBNetV2 variants. To further improve the performance, the team designed a label-matching strategy with a balanced focus on small defects and a mask voting strategy to effectively combine model predictions.
\end{description}

\noindent \textbf{General Insights}
\begin{itemize}
    \item Most top-performing teams used ensemble models, which combine the strengths of different models. 
    \item Adapting models to high-resolution input images was another important strategy, given the nature of the datasets.
    \item Data augmentation strategies were also common, and the cut-paste technique was applied by many participants. However, the effectiveness of different techniques can depend on the specific dataset.
    \item In terms of model selection, the Swin Transformer was a popular backbone choice. State-of-the-art models such as EVA and MaskDino showed strong transfer performance on the datasets.
    \item Many teams used Test Time Augmentation (TTA) to improve model predictions by averaging results over multiple augmented versions of each test image.
    \item Finally, training on all 14 datasets simultaneously was a beneficial approach for two of the top three teams, highlighting the potential benefits of multitask learning.
\end{itemize}

\subsection{Track 2}

\begin{description}
    \item[1st Place:] This solution incorporates two models, StyleGANv2 \cite{karras2020analyzing} and ControlNet \cite{zhang2023adding}, both pre-trained on large-scale datasets and subsequently fine-tuned on the training set to generate synthetic defects. The ablation study demonstrates an increase in the mean Average Precision (mAP) of downstream detection model when combining StyleGAN generated data with ControlNet generated data. It is noteworthy that traditional augmentation methods such as rotation and reflective padding are also applied during the generation process.
    \item[2nd Place:] The methodology employed consists of a data augmentation module (DAM), a data generation module (DGM), and a data selection module (DSM). Initially, the images from the training set are processed through the DAM, followed by the DGM to yield a diverse dataset. Finally, a Bayesian optimization procedure is implemented to iteratively select beneficial samples from the original, augmented, and generated datasets based on feedback signal from the validation set.
    \item[3rd Place:] The approach makes use of StyleGAN2-ADA \cite{karras2020training} to generate normal, defect-free samples and DM-CP (Cut-Paste \cite{dwibedi2017cut} combined with defect augmentation) to create new defective samples. The ablation study shows that the mAP increases with the addition of StyleGAN-ADA, and further improves when using DM-CP.
\end{description}

\noindent \textbf{General Insights}
\begin{itemize}
    \item Traditional data augmentation strategies are used in all top 3 solutions and performance improvement are observed.
    \item Generative models, i.e., GANs and Diffusion models, are popular methods for generating synthetic defects/masks/defect-free backgrounds. 
    \item The small training set size can be a problem for training Generative models. A serious discussion on the additional benefits of adopting generative models compared to traditional augmentation methods is still needed.
\end{itemize}

We will host reports from the winning teams on our workshop website\footnote{\url{https://vision-based-industrial-inspection.github.io/cvpr-2023/}} as well. Please refer to the technical reports for more technical details.

\section{Potential Research Topics}
With the VISION Datasets and Challenges, we hope to enable research in the following areas:

\noindent \textbf{Algorithms for working under Data and Annotation Limitations}
\begin{itemize}
    \item Self-supervised, semi-supervised learning, few-shot learning, and weak supervision methods for applications with rich data but limited labels.
    \item Transfer learning approaches for developing generalizable representations that can be tailored to many industrial applications with limited labeled data. The encouraging observation of positive transfer across datasets made by competition participants is a valuable first step in this endeavor.
\end{itemize}

\noindent \textbf{Data Generation Techniques that can counteract real-world data challenges}
\begin{itemize}
    \item Data-driven generative modeling techniques such as VAE, GAN, and Diffusion models.
    \item Model-driven techniques such as 3D rendering-based synthetics pipelines.
    \item Inverse rendering techniques to generate novel defect images from real-world defect images.
\end{itemize}

\noindent \textbf{Data-centric Tools and Methodologies for more suitable data collection procedures, higher data quality, and efficiency}
\begin{itemize}
    \item Data curation techniques including active learning for identifying valuable examples to label and semi-supervised learning for label quality assessment.
    \item Automatic labeling tools aiding the labeling process of human annotators.
    \item Assessment methods that can gauge the quality of datasets and annotations. 
\end{itemize}

\section{Limitations and Future Work} \label{sec:limitations}
Despite the advancements and contributions of the VISION Datasets, we acknowledge that they are not without limitations, reflecting the inherent complexity of vision-based industrial inspection tasks.

\textbf{Variability in Real-world Conditions:} While the VISION Datasets cover a wide array of manufacturing processes, materials, and industries, they cannot exhaustively replicate all possible variations encountered in real-world industrial settings. These include diverse lighting conditions, camera angles, scale, and quality of inspection images, among others. The datasets may therefore have limited applicability in scenarios that significantly deviate from the captured conditions.

\textbf{Scope of Defect Types:} Although the VISION Datasets encompass 44 distinct defect types, the exhaustive list of potential defect types in industrial inspection is beyond reach. Consequently, models trained solely on these datasets may not generalize well to detect uncommon or unrepresented defects.

\textbf{Annotation Styles:} While instance-segmentation labeling used in VISION Datasets provides a unique advantage over semantic segmentation, it may also introduce its own challenges. For instance, the boundary of certain defects might be subjective, leading to potential inconsistencies in the annotations.

Looking forward, we encourage future research to address these limitations. For instance, more comprehensive datasets could be developed to include a wider variety of real-world conditions, such as varied lighting and camera perspectives, and a broader spectrum of defect types. Additionally, further investigation could be pursued on enhancing annotation styles and methodologies to reduce potential inconsistencies and improve model performance. It would also be promising to investigate transfer learning approaches or domain adaptation techniques to improve the generalizability of models trained on the VISION Datasets to new defect types and industrial inspection scenarios.

By acknowledging and addressing these limitations, we hope to stimulate continued improvement in vision-based industrial inspection and contribute to narrowing the gap between academic research and real-world industrial practices.

\section{Conclusion}
In conclusion, the VISION Datasets aims to address the gap between academic research and industrial practice by providing a comprehensive benchmark for vision-based industrial inspection. The datasets aim to gather challenges from the broader manufacturing community and communicate them to researchers. The datasets' contributions include precise defect localization and segmentation to facilitate further decision-making in industrial practice. By enabling research in the areas of data availability, data quality, and vision-based inspection system requirements, the VISION Datasets hope to mitigate the challenges encountered in industrial practice and facilitate the development of vision-based inspection algorithms that can be applied to real-world production environments. The hosted competitions allow for the evaluation of algorithms on diverse datasets that reflect the challenges of industrial practice, ultimately enabling the development of more accurate and efficient inspection systems.

\section{Acknowledgement}

We would like to express our gratitude to Siawpeng Er and Yinwei Zhang for volunteering to be part of the dataset screening efforts.

\newpage
\bibliographystyle{unsrt}{\small
\bibliography{egbib}
}
\newpage
\appendix
\newpage
\section{Appendix}\label{sec:appendix}

\subsection{Dataset Information} \label{sec:dataset_info}

\renewcommand{\arraystretch}{1.2} 
\begin{longtable}{llccc}
    \hline
    \multirow{2}{*}{\textbf{Dataset}} & \multirow{2}{*}{\textbf{Defect}} & \multicolumn{3}{c}{\textbf{Split}} \\ 
     & & \textbf{Train} & \textbf{Validation} & \textbf{Test} \\ \hline
    \endhead
    \multirow{2}{*}{\href{https://universe.roboflow.com/roboflow-100/cable-damage}{Cable}} & Break & 33 & 95 & 98 \\ 
    & Thunderbolt & 34 & 89 & 105 \\ \hline
    \multirow{1}{*}{\href{https://universe.roboflow.com/yolotococo-kd5ab/surface-hole}{Capacitor}} & Defect & 35 & 43 & 121 \\ \hline
    \multirow{2}{*}{\href{https://universe.roboflow.com/ruben-8bqya/aughmanity_v3.0}{Casting}} & Inclusoes & 30 & 30 & 140 \\ 
    & Rechupe & 29 & 27 & 22 \\ \hline
    \multirow{4}{*}{\href{https://universe.roboflow.com/ntust-4pgai}{Console}} & Collision & 30 & 30 & 140 \\ 
    & Dirty & 30 & 30 & 139 \\ 
    & Gap & 30 & 30 & 134 \\ 
    & Scratch & 30 & 30 & 140 \\ \hline
    \multirow{4}{*}{\href{https://universe.roboflow.com/s3-vision-4/v4completev4}{Cylinder}} & Chip & 30 & 36 & 134 \\ 
    & Pistonmiss & 33 & 39 & 128 \\ 
    & Porosity & 32 & 30 & 138 \\ 
    & Rcs & 55 & 51 & 69 \\ \hline
    \multirow{1}{*}{\href{https://universe.roboflow.com/degfeg-tjrkc/qewrreq}{Electronics}} & Damage & 63 & 63 & 312 \\ \hline
    \multirow{2}{*}{\href{https://universe.roboflow.com/dataset-wzphc/hbb-inference-ori}{Groove}} & S Burr & 30 & 30 & 119 \\ 
    & S Scratch & 30 & 30 & 79 \\ \hline
    \multirow{4}{*}{\href{https://universe.roboflow.com/andy2/kodiak_defect_iv_train}{Hemisphere}} & Defect-A & 63 & 168 & 294 \\ 
    & Defect-B & 101 & 153 & 349 \\ 
    & Defect-C & 22 & 47 & 56 \\ 
    & Defect-D & 169 & 315 & 667 \\ \hline
    \multirow{5}{*}{\href{https://universe.roboflow.com/new-workspace-8oprg/contact-lens-inspection}{Lens}} & Fiber & 252 & 263 & 431 \\ 
    & Flash Particle & 8 & 8 & 6 \\ 
    & Hole & 14 & 14 & 28 \\ 
    & Surface Damage & 49 & 56 & 103 \\ 
    & Tear & 16 & 19 & 20 \\ \hline
    \multirow{6}{*}{\href{https://github.com/Ironbrotherstyle/PCB-DATASET}{PCB\_1} \cite{https://doi.org/10.1049/joe.2019.1183}} & Missing Hole & 58 & 66 & 72 \\ 
    & Mouse Bite & 39 & 36 & 35 \\ 
    & Open Circuit & 28 & 29 & 26 \\ 
    & Short & 28 & 19 & 21 \\ 
    & Spur & 23 & 15 & 15 \\ 
    & Spurious Copper & 29 & 26 & 28 \\ \hline
    \multirow{7}{*}{\href{https://universe.roboflow.com/uni-4sdfm/pcb-defects}{PCB\_2}} & Defect1 & 75 & 86 & 122 \\ 
    & Defect2 & 47 & 65 & 86 \\ 
    & Defect3 & 41 & 55 & 84 \\ 
    & Defect4 & 28 & 43 & 81 \\ 
    & Defect5 & 60 & 59 & 139 \\ 
    & Defect6 & 22 & 38 & 103 \\ 
    & Defect7 & 27 & 37 & 91 \\ \hline
\end{longtable}
    
\newpage 

\begin{longtable}{llccc}
    \hline
    \multirow{2}{*}{\textbf{Dataset}} & \multirow{2}{*}{\textbf{Defect}} & \multicolumn{3}{c}{\textbf{Split}} \\ 
     & & \textbf{Train} & \textbf{Validation} & \textbf{Test} \\ \hline
    \endhead
    \multirow{3}{*}{\href{https://universe.roboflow.com/dataset-wzphc/hbb-inference-ori}{Ring}} & T Contamination & 15 & 23 & 14 \\ 
    & T Scratch & 30 & 30 & 140 \\ 
    & Unfinished Surface & 12 & 6 & 7 \\ \hline
    \multirow{1}{*}{\href{https://universe.roboflow.com/ragnar-lodbrok-reqrw/yolov7-all-photo}{Screw}} & Defect & 75 & 75 & 150 \\ \hline
    \multirow{2}{*}{\href{https://universe.roboflow.com/1160784315-qq-com/wood-board-inspection}{Wood}} & Impurities & 42 & 53 & 135 \\ 
    & Pits & 32 & 27 & 43 \\ \hline
    \caption{Statistics of the number of instance segmentation labels we provide for the 14 VISION datasets. The datasets span various domains with different types of defects for each domain. Each dataset name is a clickable hyperlink, redirecting to the source of the dataset. Please follow the links to find the license terms of the original datasets if you would like to use them.} \\
\end{longtable}

\subsection{License} \label{sec:license}
The provided polygon annotations are licensed under \href{https://creativecommons.org/licenses/by-nc/4.0/}{CC BY-NC 4.0} License. All the original dataset assets are under the original dataset licenses.

\subsection{Disclaimer} \label{sec:disclaimer}
While we believe the terms of the original datasets permit our use and publication herein, we do not make any representations as to the license terms of the original dataset. Please follow the license terms of such datasets if you would like to use them.

\end{document}